\documentclass[lettersize,journal,compsoc]{IEEEtran}
\usepackage{amsmath,amsfonts}
\usepackage{algorithmic}
\usepackage{algorithm}
\usepackage{array}
\usepackage{amsmath,amssymb,amsfonts}
\usepackage{algorithmic}
\usepackage{graphicx}
\usepackage{textcomp}
\usepackage{tabularx}
\usepackage{multirow}
\usepackage{booktabs}
\usepackage{colortbl}
\usepackage[table ]{ xcolor}
\usepackage[caption=false]{subfig}
\usepackage{textcomp}
\usepackage{stfloats}
\usepackage{url}
\usepackage{verbatim}
\usepackage{graphicx}
\usepackage{cite}

\begin{document}

\title{
Tensor Polynomial Additive Model\\
}

\author{Yang Chen, Ce Zhu, \IEEEmembership{Fellow, IEEE}, Jiani Liu, Yipeng Liu, \IEEEmembership{Senior Member, IEEE} 
\thanks{This research is supported by the National Natural Science Foundation of China (NSFC, No. 62171088, 62020106011). Yipeng Liu is the corresponding author.}
\thanks{All the authors are with School of Information and Communication Engineering, University of Electronic Science and Technology of China (UESTC), Chengdu, 611731, China. (e-mail: yangchen2023@std.uestc.edu.cn, jianiliu@std.uestc.edu.cn, eczhu@uestc.edu.cn, yipengliu@uestc.edu.cn).}}



\IEEEtitleabstractindextext{\begin{abstract}

Additive models can be used for interpretable machine learning for their clarity and simplicity. However, In the classical models for high-order data, the vectorization operation disrupts the data structure, which may lead to degenerated accuracy and increased computational complexity. To deal with these problems, we propose the tensor polynomial addition model (TPAM). It retains the multidimensional structure information of high-order inputs with tensor representation. The model parameter compression is achieved using a hierarchical and low-order symmetric tensor approximation. In this way, complex high-order feature interactions can be captured with fewer parameters. Moreover, The TPAM preserves the inherent interpretability of additive models, facilitating transparent decision-making and the extraction of meaningful feature values. Additionally, leveraging TPAM's transparency and ability to handle higher-order features, it is used as a post-processing module for other interpretation models by introducing two variants for class activation maps. Experimental results on a series of datasets demonstrate that TPAM can enhance accuracy by up to 30\%, and compression rate by up to 5 times, while maintaining a good interpretability.


\end{abstract}

}

\maketitle

\begin{figure*}[!t]
    \centering
    \subfloat[Interpretable model performance. \label{fig:Self-interpretation model performance.}]
    {\includegraphics[width=0.27\linewidth]{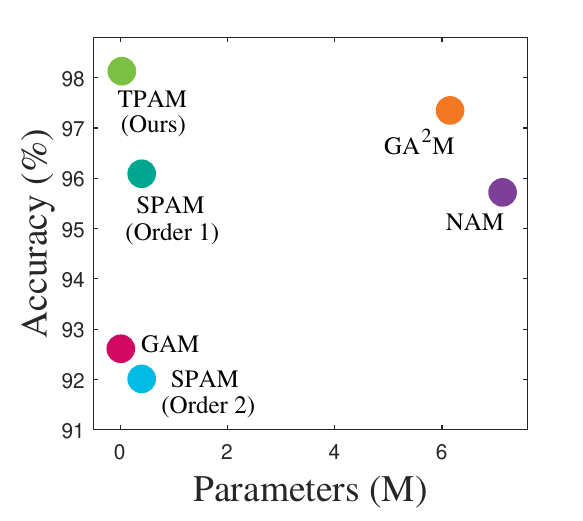}}
    \subfloat[Explanation of TPAM. \label{fig:Explanation_of_TPAM}]
    {\includegraphics[width=0.22\linewidth]{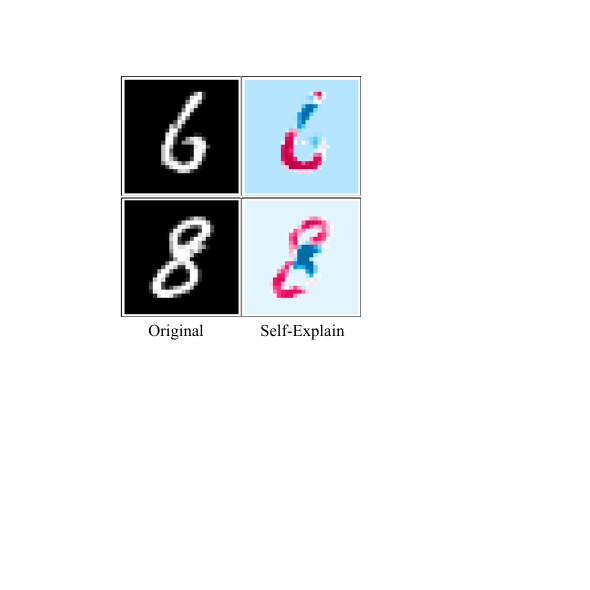}}
    \subfloat[P-CAM and PI-CAM Performance. \label{fig:P-CAM_and_PI-CAM_Performance}]
    {\includegraphics[width=0.5\linewidth]{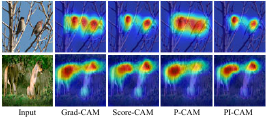}}
\caption{(a) Classification performance of the self-interpretation model on the MNIST dataset. TPAM has the highest performance with a low number of parameters. (b) Explanation of the generation of the TPAM. Red represents the positive contribution and blue represents the negative contribution, just like in subjective human judgment. (c) P-CAM and PI-CAM Performance. The resulting saliency maps are more fine-grained.}
\label{Consolidated results}
\end{figure*}
\section{Introduction}

Interpretability is crucial for understanding and validating the internal workings of machine learning models, facilitating trust, fairness, and accountability in decision-making processes. Many interpretable methods have been proposed to unveil the decision-making process of deep networks \cite{Guidotti2019Survey, Bodria2023Benchmarking, Chang2022NODEGAM}. The post-hoc interpretability techniques approximate the decisions made by an already trained model to gain insights into its behaviors \cite{Li2022Interpretable, Lundberg2017Unified, Petsiuk2018RISE, Ribeiro2016Why, Puri2021CoFrNets, Fel2023Don}, while the self-interpreting models perform transparent and understandable decision-making processes during inference without the need for additional post hoc interpretability techniques\cite{Hastie1990Generalized, Zhuang2020Interpretable, Lou2013Accurate, Agarwal2021Neural, Dubey2022Scalable}.

The generalized additive model (GAM) is a well-established approach for its straightforward interpretability \cite{Hastie1990Generalized}. To handle nonlinearly transformed input features, various GAM extensions have been developed, such as NAM \cite{Agarwal2021Neural} and GA$^2$Ms \cite{Lou2013Accurate}. Although additive models are interpretable, capturing higher-order feature interactions requires significant computational resources and leads to an exponential growth in weight space. 
The scalable polynomial additive model (SPAM) \cite{Dubey2022Scalable} utilizes polynomial approximation to improve model fitting performance, and employs tensor decomposition \cite{Dubey2022Scalable, Nie2017Generating, Brachat2010Symmetric} to reduce model complexity, thereby achieves scalable additive models. 

However, when it comes to higher-order tensor inputs, additive models based on vector representation may suffer from data structure loss and higher computational burden due to the vectorization operations, thus degenerating the system performance. 
Consequently, it is necessary to develop efficient self-interpretation models suitable for higher-order tensor data inputs.

As a transparent ``white-box" model, the additive model can not only inherently explain its own internal decisions, but also frequently embedded in other neural network methods to gain insights into their behaviors. For instance, GAMs are frequently employed as the last layer in a convolutional neural network (CNN), functioning as a component of post-hoc interpretative method. The class activation maps (CAM) \cite{Selvaraju2017GradCAM, Omeiza2019Smooth, chattopadhay2018grad, zeiler2014visualizing, springenberg2014striving, Desai2020AblationCAM, Dabkowski2017Real, wang2020ss, Fong2019Understanding} generate saliency maps by applying the weights from the classifier header to the feature maps, thereby visualizing important regions in the input image for model classification decisions. However, existing additive model classifier heads provide only a single weight, which may result in poorly fine-grained and inaccurate saliency maps \cite{ Fong2017Interpretable, Schulz2020Restricting, Wang2020ScoreCAM, BanyMuhammad2020EigenCAM}.

In this paper, we propose the tensor polynomial addition model (TPAM), which is characterized by its better accuracy, compression, and interpretation. For accuracy, TPAM directly employs high-order tensors as inputs to avoid performance loss due to vectorization. This is particularly beneficial for visual data, where TPAM enhances accuracy by integrating the a priori knowledge of CNNs with both local and global information. For compression, TPAM utilizes a hierarchical low-order symmetric tensor approximation in the weight space. It reduces computational complexity and enables the capturing of higher-order feature interactions with fewer parameters. As shown in Fig. \ref{fig:Self-interpretation model performance.}, TPAM achieves high performance with a minimal number of parameters. Regarding interpretability, as depicted in Fig. \ref{fig:Explanation_of_TPAM}, TPAM's clear decision-making process allows for extracting important values for each feature, which demonstrates its excellent self-explanation capability. Additionally, TPAM can serve as a post-processing module for other interpretative models. By integrating TPAM as a classification header in CAM, we develop two variants, i.e., P-CAM and PI-CAM. They directly input the feature map and apply a factor weighting tensor, matching the feature map's size, to achieve a more comprehensive interpretation. As shown in Fig. \ref{fig:P-CAM_and_PI-CAM_Performance}, the structured weighting of feature maps in TPAM-based P-CAM and PI-CAM significantly refines the granularity of the resulting saliency maps. 

Our main contributions can be summarized as follows:
\begin{itemize}
\item We propose an interpretable polynomial model, referred to as TPAM, which takes tensor representation for higher-order input and utilizes low-rank symmetric tensor approximation to avoid the computational complexity of high-dimensional feature interactions.

\item In the content of CAM, we propose two new methods, namely P-CAM and PI-CAM, to improve the fine-grainness of saliency maps.
P-CAM and PI-CAM replace the classification head with TPAM in CAM, enabling a structured weighting of the feature maps and thereby benefiting the generation of detailed and precise explanations.

\item As shown in Fig. \ref{Consolidated results}, the proposed TPAM achieves a high level of accuracy with a reduced number of parameters. Its capability to handle high-order inputs and transparency benefits its interpretability and the application as a post-hoc interpretation technique.

\end{itemize}

\section{Notations and Related Works}

\subsection{Notations}

 Vectors, matrices, and tensors are denoted as bold lowercase, bold uppercase, and bold Eulerian letters, e.g., $\mathbf{x}$,  $\mathbf{X}$, ${\mathcal{X}}$.
${x_{ij}}$ and $x_i$ represent the $i,j$-th element of a matrix $\mathbf{X}$  and $i$-th element of a vector ${x}$, respectively. A tensor of $D$-th order is represented as $\mathcal{X}\in\mathbb{R}^{I_1\times \cdots\times I_D}$, whose $i_1,\cdots,i_D$-th element is denoted as ${x_{i_1,\cdots,i_D}}$. The Hadamard product and outer product are denoted as $\otimes$ and $ \circ $, respectively.  $\odot_k$ stands for the inner product of the repeated pattern along the $k$-th mode. 

\subsection{Interpretable Additive Modeling Machine Learning}

The generalized additive modes (GAMs)\cite{Hastie1990Generalized} are favored in academia for their straightforward structure and transparent decision-making. However, they face limitations with complex data. The GA$^2$M model addresses this by incorporating second-order interactions into GAMs, thereby enhancing their capability. Despite its improvements, GA$^2$M struggles with large datasets due to the high computational demands during retraining. An alternative approach involves using interpretable transformations in neural networks to handle nonlinear data more effectively. For instance, neural additive models (NAMs)\cite{Agarwal2021Neural} employ neural network-based nonlinear transformations on individual features, improving model fit without the need for complex high-order interactions among features. Additionally, the NODE-GAM\cite{Chang2022NODEGAM} model merges the nonlinear processing of NAMs with the advantages of decision trees, resulting in superior performance across various datasets.

scalable polynomial additive model (SPAM) \cite{Dubey2022Scalable} considers the interactions among features in a model of arbitrary order. It utilizes polynomials to capture and fit these features, resulting in improved model performance, and employs low-rank tensor decomposition to effectively reduce the number of parameters in the model. However, the above interpretable model is limited to one-dimensional inputs. When the input data consists of matrices or higher-order tensors, data need to be vectorized, resulting in the loss of spatial structure and a subsequent decrease in model performance. Our proposed model takes higher-order tensors as direct inputs and utilizes tensor decomposition techniques to greatly reduce the number of parameters. Consequently, this approach does not introduce any performance loss.

\subsection{Saliency Map Generation}

Additive models are known for their superior transparency and are often used as part of an auxiliary interpretation tool for convolutional neural network (CNN) classifiers. In this realm, the class activation map (CAM)\cite{Zhou2016Learning} technique stands out as a distinctive post-hoc interpretation method. It efficiently creates saliency maps by utilizing the weights of the additive classifier to linearly weight the feature maps. The primary objective of this approach is to underscore the image regions most relevant to the target classification. There are two principal types of CAM techniques. The first relies on gradient backpropagation, employing the gradient of the additive model to weight the feature maps. Examples include Grad-CAM \cite{Selvaraju2017GradCAM}, Grad-CAM++ \cite{chattopadhay2018grad}, XGrad-CAM \cite{fu2020axiom}, and Layer-CAM \cite{jiang2021layercam}. The second type depends on network forward reasoning, using the weights of the additive model to generate confidence scores, which in turn determine the weights for creating the saliency maps. Notable methods here are Score-CAM\cite{Wang2020ScoreCAM}, IS-CAM\cite{naidu2020cam}, and SS-CAM\cite{wang2020ss}. However, current CAM techniques, regardless of type, can only assign a single weight to the feature maps, which constrains the precision of the resulting saliency maps.

\subsection{Polynomial Neural Network Learning}

Recent studies have demonstrated the effectiveness of polynomial interactions in approximating simple neural networks and various functions. Among these, multiplicative interactions stand out as an effective polynomial interaction model. The studies in \cite{Shin1991pisigma,Li2003SigmaPiSigma, Simonyan2015Very,voutriaridis2003ridge,oh2003polynomial} offer theoretical insights into the benefits of polynomial connections, particularly for second- or third-order interactions. Moreover, \cite{Chrysos2020Pnets, Chrysos2022Deep, Chrysos2023Regularization} expands the use of the polynomial model to higher-order multiplicative interactions. Concurrently, polynomial addition\cite{Chrysos2020Pnets, Chrysos2022Deep, Chrysos2023Regularization} emerged as an innovative approach in interaction modeling. While additive models are more resource-intensive and require careful development, they are superior in improving decision-making interpretability and transparency. 


\section{Methods}

\subsection{Scalable Polynomial Additive Models (SPAM) }

Polynomial models characterize different levels of interactions between features by adjusting the degree of the polynomial. For a vector feature input $\mathbf{x}\in \mathbb{R}^{I}$, the learning of a polynomial of order $K$ is as follows:
\begin{equation}
\label{eq:2}
\operatorname{P}(\mathbf{x})=b+\sum_{k=1}^K{\mathcal{W} ^{(k)}\odot _k\mathbf{x}}
\end{equation}
where $\mathbf{\mathcal{W}} ^{(k)} \in \mathbb{R} ^{I\times \cdots\times I}$ is a $k$-th order tensor, which denotes the weights of the $k$-th order interaction of features in $\mathbf{x}$. Large input dimension $I$ and order $K$   result in a sharp increase in the weight space's dimension $\mathcal{O}\left(I^K\right)$, which makes the computation and model optimization challenging.

To simplify the model, SPAM \cite{Dubey2022Scalable} uses CP decomposition to factorize the symmetric weight tensor $\mathbf{\mathcal{W}} ^{(k)} $ as
\begin{equation}
\label{eq:3}
\mathbf{\mathcal{W}} ^{(k)}=\sum_{r=1}^R{\lambda _{kr}}\cdot \underset{k\,\,\text{times}}{\underbrace{\left( \mathbf{u}_{kr}\circ \cdots \circ \mathbf{u}_{kr} \right) }}
\end{equation}
where $\left\{\mathbf{u}_{kr}\right\}_{r=1}^R \in \mathbb{R}^{I} $ are factor vectors, $R$ represents the CP rank, and $\left\{\lambda_{kr}\right\}_{r=1}^R$ represents the wights of each rank one component. Then the polynomial \eqref{eq:2} boils down to
\begin{equation}
\label{eq:4}
\operatorname{P}(\mathbf{x})=b+\sum_{k=1}^K{\sum_{r=1}^{R_k}{\lambda _{kr}}\cdot \left. \langle \mathbf{u}_{kr},\mathbf{x} \right. \rangle ^k}
\end{equation}

which reduces the number of parameters required for estimation from $\mathcal{O}(I^K)$ to $\mathcal{O}(IKR) $. To simplify the presentation, we assume that $R_k=R$.

However, when the input data is a high-order tensor ${\mathcal{X}}\in\mathbb{R}^{I_1\times \cdots\times I_N}$, the model needs to transform the tensor ${\mathcal{X}}$ as a vector input to match the modeling. To simplify the presentation, we assume that $I_n = I, n = 1, \cdots, N $. As a result, the complexity of the model will reach $\mathcal{O} (KR\cdot{I}^{N})$.  Moreover, the vectorization operation destroys the spatial information of the original data and thus degrades the model's performance.

\begin{figure*}[tbp]
\centering
\includegraphics[width=0.95\textwidth]{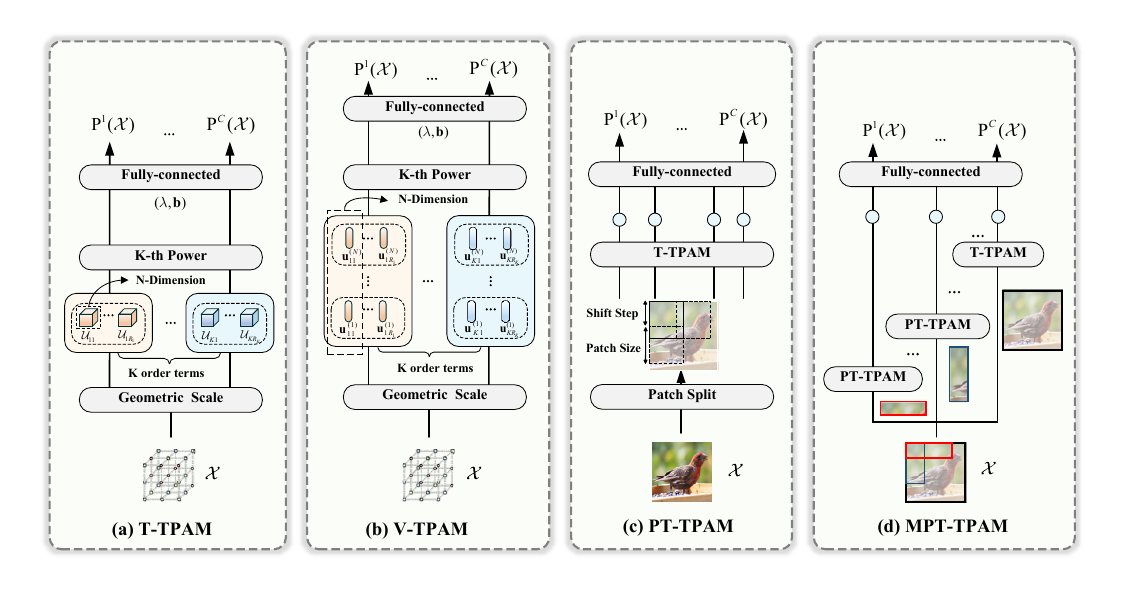} 
\caption{(a) For a $K$-order polynomial, the $K$ modules depicted in the figure correspond to different orders in the polynomial. The weight tensor $\mathbf{\mathcal{W}}$ is decomposed into a factor tensor $\mathbf{\mathcal{U}}$, which has the same size as the input $\mathbf{\mathcal{X}} \in \mathbb{R}^{I_1 \times \cdots \times I_N}$. Following inner-product and multiplication operations, $\mathbf{\mathcal{W}}$ and $\mathbf{\mathcal{U}}$ are then passed into the fully connected layer to obtain the final $C$ classification result. (b) The training process of V-TPAM remains unchanged, except for the further decomposition of $\mathbf{\mathcal{W}}$ into factor vectors $\mathbf{{u}}$. (c) PT-TPAM initially determines the patch size and shift step through patch slicing, and subsequently transfers this information to T-TPAM for further learning. (d) MPT-TPAM undergoes co-training with PT-TPAM, employing various patch sizes and shift steps. Using different patch sizes can further capture process and remote dependencies and increase the performance of the model.}
\label{fig:TPAM}
\end{figure*}

\subsection{Interpretable Tensor Polynomial Additive Model}

To reduce the number of model parameters and prevent the computational explosion problem after vectorization, a workable strategy is to maintain the tensor's original structure while discovering a new factorization rule for the weight space of feature interactions.

Specifically, we take the tensor $\mathbf{\mathcal{X}} \in \mathbb{R}^{I_1 \times \cdots \times I_N}$ directly as the feature input, and try to learn a polynomial as follows:
\begin{equation}
\label{eq:5}
\operatorname{P}(\mathcal{X})=b+\sum_{k=1}^{K}\langle\mathcal{W}^{(k)}, \underbrace{\mathcal{X} \circ \mathcal{X} \ldots \circ \mathcal{X}}_{k \text { times }}\rangle
\end{equation}
where $\mathbf{\mathcal{W}} ^{\left( k \right)}\in \mathbb{R} ^{I_1\times \dots \times I_N\times\cdots\times I_1\times \dots \times I_N }$ is a $kN$-th order weight tensor that encapsulates how feature tensors interact. We refer to this model as the Tensor Polynomial Additive Model (TPAM). 
For accuracy, we decompose $\mathbf{\mathcal{W}} ^{\left( k \right)}$ into combinations of lower order factor tensors/vectors in a hierarchical way, resulting in T-TPAM and V-TPAM, respectively. Then we employ patch splitting to capture fine-grained details and further consider training on multiple scales for global dependencies capturing. As illustrated in Fig \ref{fig:TPAM}, we present the overall structure of the TPAM.

\subsubsection{T-TPAM}

First, we decompose ${\mathcal{W}} ^{(k)}$ into a combination of several basis factor tensors of the same size as $\mathcal{X}$ as follows:
\begin{equation}
\label{eq:6}
{\mathcal{W}} ^{(k)}=\sum_{r=1}^{R_k}{\lambda _{kr}}\cdot \underset{k\,\,\text{times}}{\underbrace{\left({\mathcal{U}} _{kr}\circ \cdots \circ {\mathcal{U}}_{kr} \right)}},
\end{equation}
where $\left\{ \mathcal{U}_{kr} \in \mathbb{R} ^{I_1\times \dots \times I_N}\right\} _{r=1}^{R_k}$ is the factor tensor of ${\mathcal{W}} ^{(k)}$, $\left\{ \lambda _{kr} \right\} _{r=1}^{R_k}$ is the corresponding weight of each component. The polynomial model \eqref{eq:5} can be expressed as 
\begin{equation}
\label{eq:7}
\operatorname{P}(\mathcal{X})=b+\sum_{k=1}^K \sum_{r=1}^{R_k} \lambda_{k r} \cdot\left\langle\mathcal{U}_{k r}, \mathcal{X}\right\rangle^k.
\end{equation}

We refer to the aforementioned model as T-TPAM, which decreases the model complexity of TPAM from $\mathcal{O} (I^{NK})$ to $\mathcal{O} (KR\cdot{I}^{N})$.  As it shows in Fig.  \ref{fig:TPAM}a,  T-TPAM first applies a geometric scaling to $\mathcal{X}$, followed by $K$ polynomial modules which perform multiplication and inner product operations over $\mathcal{X}$ and $\mathcal{U}_{kr}$. Then a linear layer is used to combine polynomials with different degrees.

\subsubsection{V-TPAM}
While the TPAM effectively reduces model complexity, the size of $\mathcal{X}$ and the factor tensors $\mathcal{U}_{kr}$ can be large in practical applications, making the computational effort of T-TPAM unacceptable. Further decomposition of $\mathcal{U}_{kr}$ seems promising. 
We consider the CP decomposition as
\begin{equation}
\label{eq:8}
{\mathcal{U}}_{kr}=\sum_r^R{\mathbf{u}_{kr}^{(1)}\circ ...\circ \mathbf{u}_{kr}^{(N)}},
\end{equation}
T-TPAM can be then transformed into
\begin{equation}
\label{eq:9}
\begin{aligned}
\operatorname{P}(\mathcal{X} )=b+\sum_{k=1}^K{\sum_{r=1}^{R_k}{\lambda _{kr}}\cdot \left. \langle \mathbf{u}_{kr}^{(1)}\circ ...\circ \mathbf{u}_{kr}^{(N)},\mathcal{X} \right. \rangle ^k},
\end{aligned}
\end{equation}
where $\{\{ \mathbf{u}_{kr}^{(n)} \} _{r=1}^{R_k} \} _{n=1}^{N}$ and $ \{ \lambda _{kr} \} _{r=1}^{R_k} $ represent the learnable basic factor vectors and weights of each component for $\mathbf{\mathcal{W}}^{(k)}$.

We refer to this variant as V-TPAM, which reduces the model complexity from $\mathcal{O} (KR\cdot{I}^{N})$ to $\mathcal{O}(KR\cdot NI)$. As shown in Fig. \ref{fig:TPAM}b, the training process is similar to that of T-TPAM. However, in V-TPAM, the factor tensor $\mathcal{U}_{kr}$ is further decomposed into factor vectors $\{\mathbf{u}_{kr}^{(n)}\}_{n=1}^N$, which yields a more structured model with fewer parameters.

\subsubsection{PT-TPAM}

Weight sharing is a commonly used strategy in neural networks to promote parameter sharing and reduce the overall number of parameters in the model. Weight sharing can reduce the number of parameters in T-TPAM. To reduce model complexity, the T-TPAM is constructed solely for the local space of $\mathcal{X}$, resembling a convolution kernel, rather than modeling the entire input feature space. We denote this model using localized patches as PT-TPAM. The overall block diagram of PT-TPAM is shown in Fig. \ref{fig:TPAM}c. Factor tensor $\mathcal{U}$  with a fixed patch size and shift step is employed. During training, the tensor $\mathcal{U}$ is shifted through the entire $\mathcal{X}$ space, capturing local relationships. The linear layer is responsible for generating the final output. PT-TPAM and T-TPAM are equivalent when the patch size matches that of $\mathcal{X}$.

\subsubsection{MPT-TPAM}

Although it is with decreased model complexity, PT-TPAM has a limited receptive field, similar to that of a convolutional kernel. This limits its ability to effectively capture global information. To incorporate both local and global information, we propose to add larger patch sizes into the PT-TPAM model for a multi-training approach. The combined method is referred to as MT-TPAM, as depicted in Fig. \ref{fig:TPAM}d. In the specific training stage, we utilize a T-TPAM with a patch size identical to $\mathcal{X}$, along with multiple PT-TPAMs of smaller patch sizes for co-training.  For instance, in the case of a $32\times32$ image size, we utilize patches of sizes $4\times4$, $8\times8$, and $32\times32$ during the training.

\subsection{Learning and Discussion for TPAM}
In this section, the discussion is focused on optimizing the learning process of TPAM models, encompassing aspects such as model initialization, geometric scaling, and approaches to multi-classification challenges. Besides, this section delves into the benefits of TPAM in model compression, highlighting its efficacy and potential advantages.
\subsubsection{Initialization scheme}

Initialization is very important for neural networks. However, the current neural network regularization techniques \cite{glorot2010understanding, he2015delving} are primarily designed for convolutional and fully connected networks. Additionally, the initialization used for polynomial networks mainly considers depth\cite{Chrysos2023Regularization}. TPAM stands out as a distinctive model due to its shallow and wide polynomial structure. Consequently, the previous initialization methods are not suitable for TPAM. Based on practical experiments, it has been observed that the magnitude of higher-order weights decreases as the number of terms increases. Additionally, the weights of higher-order terms in polynomials are significantly smaller in comparison to the weights of lower-order terms. Here we propose a simple initialization scheme

\begin{equation}
\mathbf{\mathcal{U}}_{k r}, \mathbf{u}_{k n r} \sim \mathcal{N}\left(0, \sigma^2 \mathbf{I}\right), \text{where}~\sigma=Qk\sqrt{\frac{1}{M_k}},
\end{equation}
$M_k$ is the number of parameters in the $k$-th order term, $k$ is the order in which the polynomial is expanded, and $Q$ is the initialization factor.  We call this Poly-mean initialization. 

\subsubsection{Multi-classification Issues}

For multiclassification problems with $C$ classes, applying a polynomial model requires learning multiple polynomials simultaneously. It could lead to performance degradation and overfitting, especially when dealing with numerous classes.

To simplify the model, a shared factor tensor $\mathcal{U}_{kr}$ across all classes is introduced, similar to that of SPAM. This means that the same factor tensor is utilized for each class. Additionally, we use class-specific eigenvalues $\{ \lambda _{kr}^{c}\} _{c=1}^{C}$ and bias values $\{b^{c} \} _{c=1}^{C}$ to learn the unique characteristics of each class. Taking T-TPAM as an example, we investigate the following polynomials for class $c$ as follows:

\begin{equation}
\label{eq10}
\operatorname{P}^c(\mathcal{X} )=b^c+\sum_{k=1}^K{\sum_{r=1}^{R_k}{\lambda _{kr}^{c}}}\cdot \left. \langle \mathcal{U} _{kr},\mathcal{X} \right. \rangle ^k
\end{equation}

\subsubsection{Geometric rescaling}
As the order of the interactions increases, training models to learn higher-order interactions becomes challenging. Moreover, the values obtained from such interactions become disproportionately larger than the original feature values. This presents a significant challenge to both the fitting ability of TPAM and its interpretability. Similar to SPAM, to mitigate the disproportionately negative impact of higher-order interactions, we employ a geometric scaling approach to adjust the TPAM model. For $K$-order interactions with input $\mathbf{\mathcal{X}}$, scale scaling is defined as $\mathbf{\tilde{\mathcal{X}}}_K=\mathrm{sign(}\mathbf{\mathcal{X}})\cdot |\mathbf{\mathcal{X}}|^{1/K}$.

\subsubsection{Model Compression}

Here we consider the input tensor ${\mathcal{X}} \in \mathbb{R}^{I_1 \times \ldots \times I_N}$ and a polynomial function with order $K$. Table \ref{tab：Comparison of SPAM and TPAM parameters} illustrates the number of model parameters required by GAM, SPAM, T-SPAM, and V-TPAM. All four models utilize the same linear function. TPAM maintains the multiway structure, whereas, for SPAM, the input tensor must be vectorized. 

Table \ref{tab：Comparison of SPAM and TPAM parameters} demonstrates that particularly for high-order tensor data, both TPAM and SPAM have considerably fewer parameters compared with GAM. Furthermore, the accumulation of dimensions is significantly smaller than the cumulative multiplication of dimensions, i.e., $I^{N} >> NI$, 
which makes V-TPAM require far fewer parameters than both SPAM and T-TPAM and thus highly scalable and suitable for large datasets.

\begin{table}[t]
    \centering
        \renewcommand\arraystretch{1.5} 
    \small
    \caption{Complexity of GAM, SPAM, TPAM. }
    \begin{tabular}{cc}
    \hline 
    \hline
        Model & Parameters for input $\mathbf{\mathcal{X}} \in \mathbb{R}^{I_1 \times \ldots \times I_N}$ \\ 
        \hline
        {\sc GAM}     & $I^{NK}$ \\
        {\sc SPAM}    & $KR\cdot I^{N}$ \\
        {\sc T-TPAM}  & $KR\cdot I^{N}$ \\
        {\sc V-TPAM}  & $KR\cdot NI$ \\ \hline
        \hline
    \end{tabular}
    \label{tab：Comparison of SPAM and TPAM parameters}
\end{table}

\section{Model Interpretation}

TPAM demonstrates transparency in its internal decision-making processes and exhibits a high level of interpretability. In this section, our focus lies on examining the interpretability of the TPAM model itself, as well as its application as a post-hoc module in the context of CAM.
\begin{figure*}[t]
\centering
\includegraphics[width=0.9\textwidth]{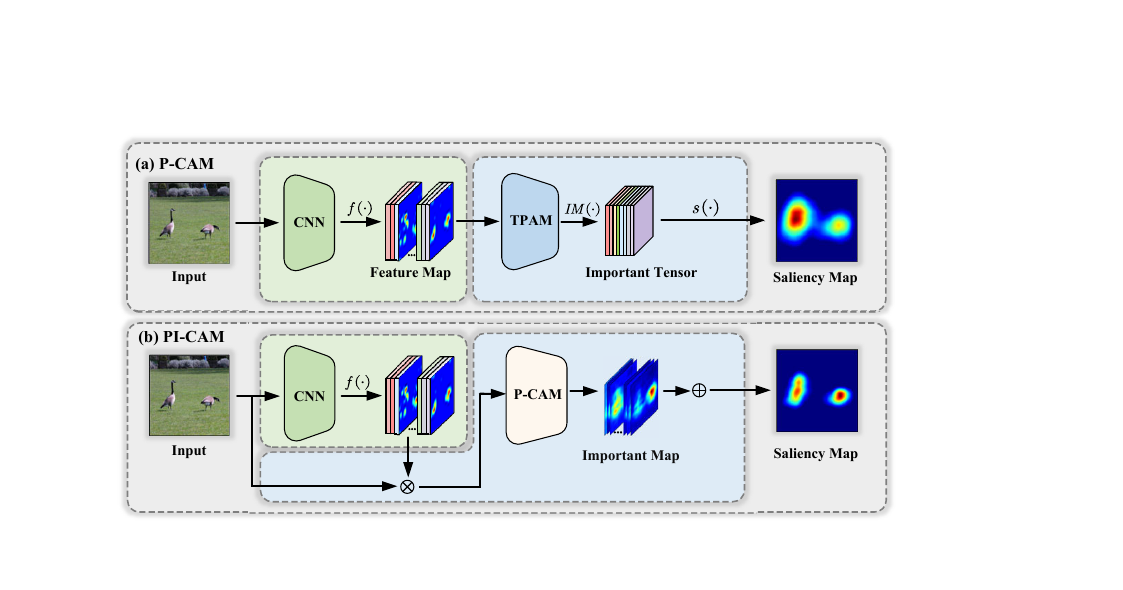} 
\caption{(a) In P-CAM, the input $\mathbf{\mathcal{X}}$ undergoes processing through CNN model $f(\cdot)$, resulting in the acquisition of feature map $\mathbf{\mathcal{A}}_l$. Subsequently, the TPAM model classifies the feature map, and the trained TPAM is employed to extract the model's important tensor. The saliency map is then obtained by downscaling the important tensor using the function $s(\cdot)$. (b) PI-CAM acquires the feature map $\mathbf{\mathcal{A}}_l$ through the CNN model $f(\cdot)$. Feature map is then utilized as input to P-CAM after performing a dot product with the input $\mathbf{\mathcal{X}}$. Subsequently, P-CAM is trained to obtain the important map (the saliency map of each $\mathbf{{A}}_l^{i_c}$). The saliency map is derived by adding the important maps.
}
\label{fig:P-CAM and PI-CAM}
\end{figure*}

\subsection{Self-Explanation}

The additive model boasts a straightforward structure where the contributions of individual features combine linearly to form the final output.
The interaction between features and weights determines the importance of each feature in the model. 
Positive values indicate positive contributions, while negative values have negative effects. For example, for the GAM (Linear) model \cite{Hastie1990Generalized}, its importance for the input $\boldsymbol{x}=\left\{ x_1,...,x_N \right\}$  can be expressed as $w_i\cdot x_i,i\in \{1,...,N\}$, $w_i$ is the weight value corresponding to $x_i$. Extracting the importance of each feature and organizing them in the same spatial location as the input, we can get a corresponding important map.

For a $K$-th order SPAM \cite{Dubey2022Scalable} model, the feature important vector is 
\begin{equation}
\label{eq11}
\operatorname{IM}_{\operatorname{SPAM}}\left( \mathbf{x} \right) =\sum_{k=1}^K{\sum_{r=1}^{R_k}{\lambda_{kr} \cdot\left( \mathbf{u}_{kr}\otimes \mathbf{x} \right) \cdot \left. \langle \right. \mathbf{u}_{kr},\mathbf{x}\rangle ^{k-1}}}
\end{equation}
where $\mathbf{x}=\operatorname{vec}(\mathcal{X})$, $\mathbf{u}_{kr}$ is the weight factor vector capturing the interactions of $\mathbf{{x}}$.
For TPAM, the feature important tensor, denoted as $\operatorname{IM}_{\operatorname{TPAM}}$, can be defined as 
\begin{equation}
\label{eq12}
\operatorname{IM}_{\operatorname{TPAM}}\left( \mathbf{\mathcal{X}} \right) =\sum_{k=1}^K{\sum_{r=1}^{R_k}{\lambda_{kr} \cdot\left(  \mathbf{\mathcal{U}} _{kr}\otimes \mathbf{\mathcal{X}} \right) \cdot \left. \langle \right. \mathbf{\mathcal{U}} _{kr},\mathbf{\mathcal{X}} \rangle ^{k-1}}}
\end{equation}
where interactions of ${\mathcal{X}}$ are represented by the weight factors $\mathbf{\mathcal{U}} _{kr}$.

\begin{figure}[tbp]
\centering
\includegraphics[width=0.48\textwidth]{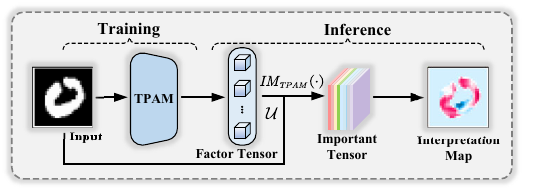}
\caption{General process of self-interpretation in TPAM. }
\label{fig:General process of self-interpretation}
\end{figure}

 This important vector/tensor is then visualized to generate intuitive interpretations. The details of this process are illustrated in Fig. \ref{fig:General process of self-interpretation}. In the case of image data, we often sum the important tensor along the channel dimension, resulting in an important matrix commonly referred to as an interpretation map. For instance, the region that TPAM concentrates on is visible in the explanation map produced on MNIST\cite{xiao2017fashion}, which is essentially the same as human subjective judgment, as shown in Fig. \ref{fig:General process of self-interpretation}. The model performance improvement can also be partially explained by the generated explanation map because it is a direct result of the weights within the model.

\subsection{P-CAM \& PI-CAM}

TPAM offers superior performance and internal decision-making transparency, making it suitable as a classifier instead of a conventional linear layer in various neural networks. For instance, in the case of CNNs, TPAM can be post-positioned on the last feature layer, similar to CAM\cite{Zhou2016Learning}, to extract the importance of that layer and generate reliable saliency maps without compromising performance. Furthermore, CAM and Score-CAM\cite{Wang2020ScoreCAM} often produce fine-grained maps with subpar accuracy due to the single weight of the feature layers. In contrast, TPAM can weigh the feature maps using weights of the same size, resulting in more accurate and finer-grained saliency maps. To achieve this, we propose the P-CAM and PI-CAM algorithms for CNNs and image inputs.

\subsubsection{P-CAM} 

Given input data $\mathbf{\mathcal{X}} $, we use the network action $\operatorname{f}(\cdot)$ to interpret the CNN network. We pick an internal convolutional layer $l$ in $\operatorname{f}$ and represent the corresponding feature map as $\mathbf{\mathcal{A}}_l=\operatorname{f}(\mathbf{\mathcal{X}}) \in \mathbb{R}^{I_C \times I_H \times I_W}$. Then the P-CAM saliency map for class $c$ is specified as $\operatorname{L}_{\text{P-CAM}}^{c}$ 
\begin{equation}
L_{\operatorname{P-CAM}}^c={\operatorname{ReLU}}\left(\operatorname{norm}\left(\operatorname{s}\left(\mathbf{\alpha}^c_l \right)\right)\right)
\end{equation}
where
\begin{equation}
\mathbf{\mathcal{\alpha}}^c_l=\operatorname{IM}^c\left(\mathbf{\mathcal{A}}_l\right)\in \mathbb{R}^{I_C \times I_H \times I_W}
\end{equation}
is the important tensor of $\mathbf{\mathcal{A}}_l$, $\operatorname{s}(\cdot): \mathbb{R}^{I_C \times I_H \times I_W} \rightarrow \mathbb{R}^{I_H \times I_W}$ calculates the total sum over the channel dimension, $\operatorname{norm}(\cdot)$ represents the normalization operation, and $\operatorname{IM}(\cdot)$ illustrates the way to extract important tensor/vector as defined in \eqref{eq11} and \eqref{eq12}. In Fig. \ref{fig:P-CAM and PI-CAM}a, the overall flow of P-CAM is depicted. After obtaining the saliency map, the ReLU operation is applied due to its experimental effectiveness.  

\subsubsection{PI-CAM}

Standard CAM methods typically apply the weights from additive models directly onto feature maps. However, this approach overlooks the interactions between feature maps, which can compromise the integrity of the weight extraction process in these models. Score-CAM attempted to rectify this issue by incorporating confidence scores. Yet, its reliance on a singular weight for feature map weighting tends to produce saliency maps that are both unclear and imprecise.

In response, we introduce the PI-CAM algorithm, an enhancement of P-CAM and Score-CAM methodologies.  It achieves this by placing the TPAM after the final convolutional layer in the CNN and employing pixel-level weights to weight the feature maps. This method ensures a more accurate and visually fine-grained of the saliency maps.

Specifically, for a class $c$, PI-CAM generates a saliency map as 
\begin{equation}
\operatorname{L}_{\operatorname{PI-CAM}}^c={\operatorname{ReLU}}\left(\operatorname{norm}\left(\sum_{i_c}^{I_C} s\left(\mathbf{\alpha}_{i_c=1}^c\right)\right)\right)
\end{equation}
where
\begin{equation}
\mathbf{\alpha}_{i_e}^c=\operatorname{IM}^c\left(\operatorname{f}\left(\mathbf{\mathcal{X}} \otimes \mathbf{\mathcal{H}}_l^{i_c}\right)\right)-\operatorname{IM}^c\left(\operatorname{f}\left(\mathbf{\mathcal{X}}_b\right)\right)
\end{equation}
with
\begin{equation}
\mathbf{\mathcal{H}}_l^{i_c}=\operatorname{norm}\left(\operatorname{\phi}\left(\mathbf{A}_{l}^{i_c}\right)\right)
\end{equation}
$\mathbf{A}_l^{i_c} \in \mathbb{R}^{I_H \times I_W}$ is the $i_c$-th channel of $\mathbf{\mathcal{A}}_l$, and $\operatorname{\phi}(\cdot): \mathbb{R}^{I_H \times I_W} \rightarrow \mathbb{R}^{I_C \times I_H \times I_W}$ denotes the operation that upsamples $\mathbf{A}_{l}^{i_c}$ into the same size of the input. $\mathbf{\mathcal{X}}$ represents the image requiring interpretation, while $\mathbf{\mathcal{X}}_b$ serves as the baseline image, typically consisting of either all white or all black.

The pipeline of the proposed PI-CAM is illustrated in Fig. \ref{fig:P-CAM and PI-CAM}b. 
In PI-CAM, images undergo CNN processing to generate feature maps. These feature maps from various channels are then element-wise multiplied with the input image. The multiplication outcomes, along with the input image, act as inputs to the P-CAM module. P-CAM produces important maps for each channel, and these maps are summed to produce the final saliency map.

\begin{figure}[t]
\centering
\includegraphics[width=0.48\textwidth]{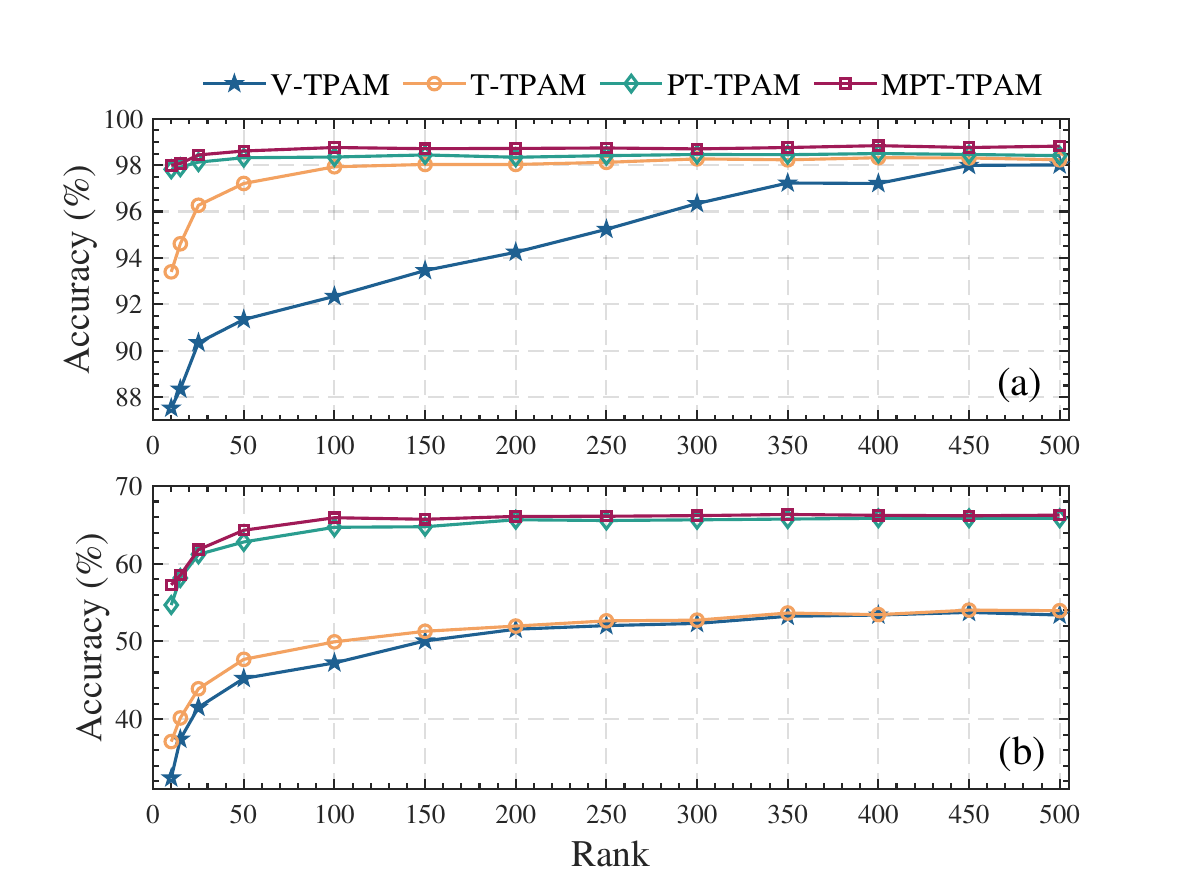} 
\caption{The effect of different TPAM models on model performance is tested on different datasets for Ranks. (a) MNIST. (b) CIFAR-10}
\label{fig:rank}
\end{figure}

\section{Experiments}

In the forthcoming sections, we conduct a thorough evaluation of TPAM's performance, focusing on its superior outcomes and notable interpretability, through various experiments. Initially, the section \ref{Performance Evaluation of TPAM Classification} examines the impact of parameter settings and model initialization on TPAM's accuracy. We then compare TPAM with various established, interpretable baseline models to showcase its enhanced performance across datasets of different sizes. Following this, In section \ref{Interpretation Evaluation}  TPAM is compared with several other post hoc interpretable models. This comparison aims to confirm TPAM's self-interpretation abilities and to investigate the reasons for its exceptional classification task performance. The paper also assesses the effectiveness of TPAM-derived PI-CAM and P-CAM algorithms, further supporting our conclusions.

\subsection{Performance Evaluation of TPAM Classification}\label{Performance Evaluation of TPAM Classification}

In section \ref{Ablation Studies}, we assess the performance of TPAM by conducting an ablation study on its hyperparameters. Then we compare the proposed TPAM with existing interpretable models on several real datasets for image classification in section \ref{Measuring Benchmark Performance}. As model inputs, we use MNIST\cite{lecun2015deep}, CIFAR-10\cite{krizhevsky2009learning}, and STL-10 standard datasets and crop them to $28 \times 28 \times 1$, $32 \times 32 \times 3 $,  and $96 \times 96 \times 3 $ sizes without using data enhancement operations. For all datasets with no defined train-val-test split, we use a fixed random sample of 70\% of the data for training, 20\% for validation, and 10\% for testing.  Each model undergoes 10 training iterations on Pytorch using the Adam\cite{kingma2014adam} optimization algorithm, with only the optimal results being reported.

\begin{figure}[t]
\centering
\includegraphics[width=0.49\textwidth]{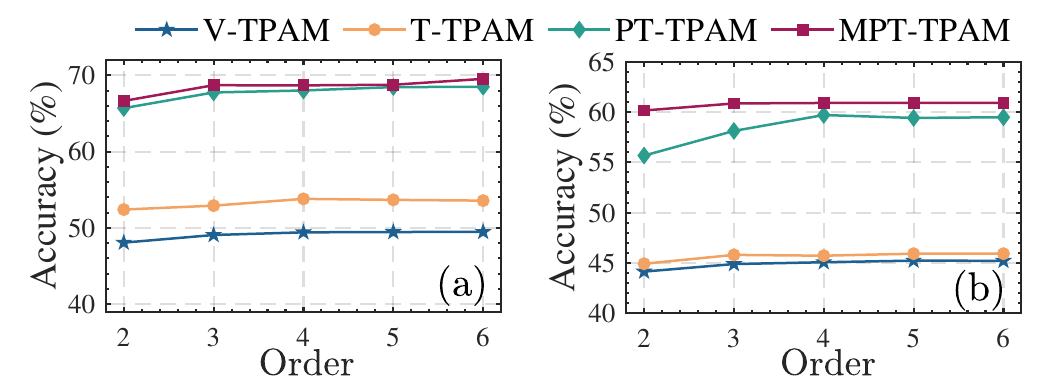} 
\caption{The effect of different TPAM models on model performance is tested on different datasets for Orders. (a) CIFAR-10. (b) STL-10. }
\label{fig:order}
\end{figure}

\subsubsection{Ablation Studies} \label{Ablation Studies}
Building upon the foundational experiments outlined previously, this section delves into a detailed analysis of TPAM's hyperparameters. Specifically, we scrutinize the influence of factors such as the order and rank, initialization scheme, patch size, and shift step on the classification performance of TPAM.

\begin{table*}[t]
	\centering
	\caption{Evaluation of TPAM on benchmarks against prior work. ($\uparrow$): the higher the better; ($\downarrow$): the lower the better. \textbf{Bold} indicates the best performance and the least parameters (excluding uninterpretable baselines). The best results were reported with optimal hyperparameters and 10 randomized trials.}
	\resizebox{\linewidth}{!}{
\begin{tabular}{lcccccccccccc}
\hline \hline
\multicolumn{1}{l}{\multirow{3}{*}{Method}} & \multicolumn{8}{c}{2D}                                                                                                                              & \multicolumn{4}{c}{3D}                                          \\
\cmidrule(lr){2-9}
\cmidrule(lr){10-13}
\multicolumn{1}{c}{}                        & \multicolumn{2}{l}{ArticularyWordRecognition} & \multicolumn{2}{c}{Heartbeat}  & \multicolumn{2}{c}{MNIST}      & \multicolumn{2}{c}{Fashion-MNIST} & \multicolumn{2}{c}{CIFAR-10}    & \multicolumn{2}{c}{STL-10}      \\
\cmidrule(lr){2-3}
\cmidrule(lr){4-5}
\cmidrule(lr){6-7}
\cmidrule(lr){8-9}
\cmidrule(lr){10-11}
\cmidrule(lr){12-13}
\multicolumn{1}{c}{}                        & Acc$\uparrow$ & Param$\downarrow$                & Acc$\uparrow$ & Param$\downarrow$         & Acc$\uparrow$ & Param$\downarrow$         & Acc$\uparrow$ & Param$\downarrow$          & Acc$\uparrow$ & Param$\downarrow$         & Acc$\uparrow$ & Param$\downarrow$         \\
\hline
\multicolumn{13}{c}{Uninterpretable   Black-Box Baselines}                                                                                                                                                                                                          \\
\hline
DNN                                         & 97.67                  & 5.30                 & 72.68          & 53.23         & 98.02          & 1.60          & 90.28            & 2.67           & 59.62          & 4.21          & 35.81          & 62.50         \\
SOTA                                       & 99.00                  & 7.22                 & 76.60          & 6.97          & 99.87          & 1.51          & 94.91            & 24.73          & 99.10          & 23.00         & 97.20          & 5.10          \\
\hline
\multicolumn{13}{c}{Interpretable   Baselines}                                                                                                                                                                                                                      \\
\hline
NAM                                         & 96.21                  & 0.06                 & 69.74          & 1.13          & 95.72          & 7.28          & 89.48            & 6.73           & 41.79          & 27.90         &    ---            & 251.54        \\
GAM (Linear)                              & 96.00                  & \textbf{0.03}        & 66.83          & \textbf{0.05} & 92.61          & \textbf{0.01} & 84.79            & \textbf{0.01}  & 36.40          & 0.03          & 30.68          & 0.27          \\
GA2M (Linear)                              & 96.35                  & 36.33                &     ---           & 1220.69       & 97.35          & 6.15          & 85.72            & 6.15           & 47.83          & 94.40         &      ---        & 7644.39       \\
SPAM (Order 1)                              & 96.00                  & 0.26                 & 70.73          & 4.95          & 92.01          & 0.40          & 84.12            & 0.32           & 40.20          & 1.54          & 30.90          & 11.06         \\
SPAM(Order 2)                               & 95.61                  & 0.28                 & 72.20          & 4.95          & 96.09          & 0.40          & 85.22            & 0.32           & 45.27          & 1.54          & 32.04          & 11.06         \\
\hline
\multicolumn{13}{c}{Our Interpretable   Models}                                                                                                                                                                                                                     \\
\hline
V-TPAM (Order 1)                            & \textbf{96.67}         & 0.04                 & 73.20          & 0.09          & 96.73          & 0.03          & 84.85            & 0.03           & 42.54          & \textbf{0.03} & 42.32          & \textbf{0.10} \\
V-TPAM (Order 2)                            & 95.42                  & 0.04                 & \textbf{74.15} & 0.09          & 98.03          & 0.03          & 90.68            & 0.03           & 52.03          & 0.03          & 46.06          & 0.10          \\
T-TPAM (Order 2)                            & 96.31                  & 0.58                 & 74.01          & 19.76         & 98.31          & 0.36          & 90.27            & 0.36           & 53.95          & 1.54          & 46.47          & 13.83         \\
PT-TPAM (Order 2)                           & \multicolumn{4}{c}{\multirow{2}{*}{------}}                                          & 98.46          & 0.85          & 91.09            & 0.85           & 65.84          & 0.94          & 59.97          & 2.59          \\
MPT-TPAM (Order 2)                          & \multicolumn{4}{c}{}                                                           & \textbf{98.73} & 1.25          & \textbf{91.24}   & 1.25           & \textbf{66.24} & 2.48          & \textbf{60.55} & 16.42       \\ 
\hline
\hline
\end{tabular}}
	\label{tab:different_inter_xperiments}
\end{table*}

\begin{table}[t]
	\centering
	\caption{Performance of TPAM with different initializations.
 }
\begin{tabular}{llll}
\hline\hline
\multicolumn{1}{c}{Model} & \multicolumn{1}{c}{Initialization} & \multicolumn{1}{c}{Order 2} & \multicolumn{1}{c}{Order 4} \\
\hline
\multirow{5}{*}{T-TPAM}   & Xavier                             & 53.54                       & 53.87                       \\
                          & Orthogonal                         & 53.59                       & 54.10                       \\
                          & Kaiming normal                     & 52.97                       & 54.04                       \\
                          & Kaiming uniform                    & 53.28                       & 54.10                       \\
                          & Poly-mean                          & \textbf{53.65}              & \textbf{54.59}              \\
                          \hline 
\multirow{5}{*}{PT-TPAM}  & Xavier                             & 64.25                       & 68.58                       \\
                          & Orthogonal                         & 64.21                       & 68.09                       \\
                          & Kaiming normal                     & 63.69                       & 66.75                       \\
                          & Kaiming uniform                    & 63.89                       & 66.43                       \\
                          & Poly-mean                          & \textbf{65.36}              & \textbf{68.95}   \\
                          \hline \hline
\end{tabular}
	\label{tab: initialisation}
\end{table}

\textbf{Order and Rank.} Earlier research on polynomial approximation has shown that both the order of polynomial expansion and the required rank for weight space reconstruction significantly influence TPAM's performance. To delve deeper into this relationship, we conducted ablation studies on the MNIST, CIFAR-10, and STL-10 datasets, specifically focusing on variations in adopted rank and order. Initially, we set the order to $2$ and varied the rank from $50$ to $500$. Results for MNIST and CIFAR-10 are illustrated in Fig. \ref{fig:rank}(a) and Fig. \ref{fig:rank}(b), respectively. Subsequently, the rank was fixed at $200$, and the order varied within the range of $\{2,3,4,5,6\}$. Results for MNIST and STL-10 are depicted in Fig. \ref{fig:order}(a) and Fig. \ref{fig:order}(b), respectively. 

As expected, TPAM's performance improves with increases in both rank and order, where PT-TPAM and MPT-TPAM consistently outperform others. Specifically, with a constant order, as shown in Fig. \ref{fig:order}a, all four TPAM models show continual performance gains with increasing rank, although the rate of improvement slows over time. Notably, V-TPAM demonstrates rapid improvement on the MNIST dataset, indicating a need for a higher rank in accurate weight space reconstruction. Furthermore, with a fixed rank, elevating the order results in enhanced model performance, albeit marginally. Therefore, lower orders suffice to achieve satisfactory classification results.

\textbf{Model Initialization.} The initialization of a model plays a crucial role in determining its performance. In our study, we employ widely recognized initialization schemes, including xavier initialization\cite{glorot2010understanding}, orthogonal initialization\cite{pennington2017resurrecting}, and Kaiming initialization\cite{he2015delving}, and our proposed initialization scheme. To evaluate their performance, we applied these techniques as the initialization module of both T-TPAM and MPT-TPAM models and conducted experiments on the MNIST dataset. The rank is set as 200, order is chosen from $\{2,4\}$.
The experimental results are exhibited in  Table \ref{tab: initialisation}.  
As anticipated, previously proposed initialization schemes are unable to align with the structure of TPAM, consequently resulting in suboptimal training outcomes.  In contrast, the multi-means initialization scheme demonstrates remarkable efficacy for TPAM. 

\begin{figure}[t]
\centering
\includegraphics[width=0.99\columnwidth]{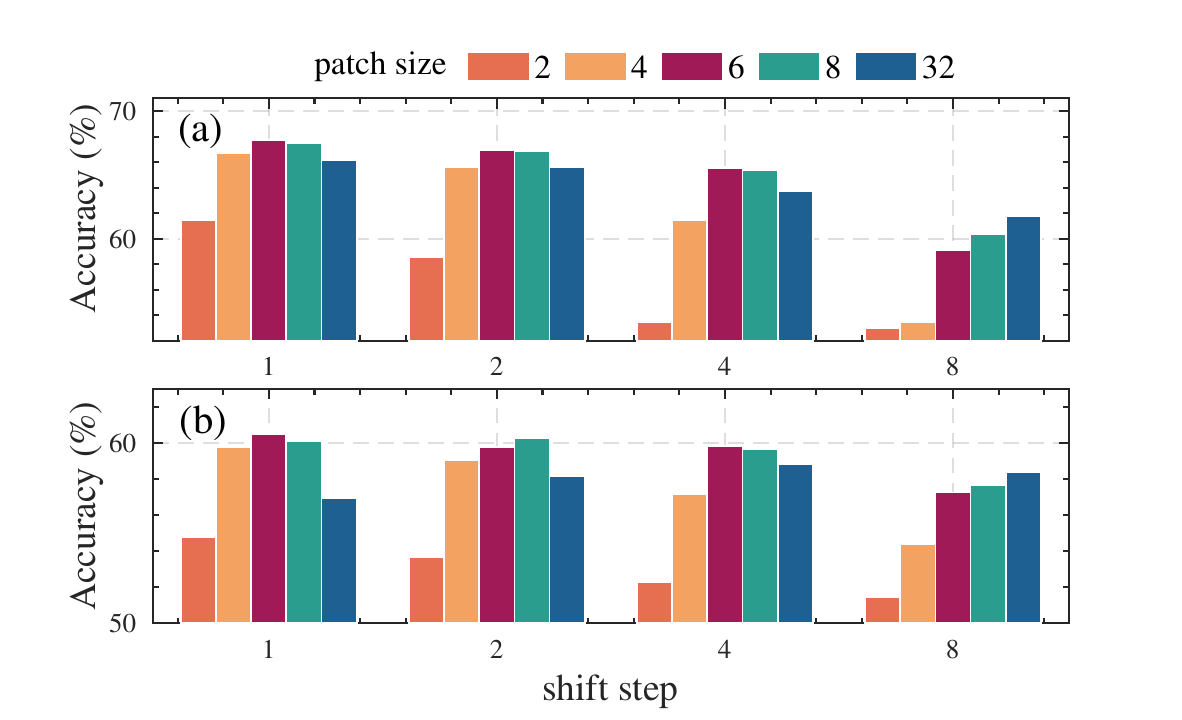} 
\caption{The impact of patch size and shift step size in PT-TPAM on the classification performance of (a) CIFAR-10 and (b) STL-10 datasets.}
\label{fig:Impact of patch size and shift step size}
\end{figure}

\textbf{Patch Size and Shift Step}. As depicted in Fig. \ref{fig:rank}, P-TPAM exhibits exceptional performance. However, PT-TPAM is significantly influenced by the shift step and the patch size. To delve deeper into this correlation, we conducted experiments on both the CIFAR-10 and STL-10 datasets with  PT-TPAM while maintaining a constant rank and order of 200 and 2, respectively.  The results are shown in Fig. \ref{fig:Impact of patch size and shift step size}. On both datasets, PT-TPAM performs well with a patch size of 8 and a shift step size of 2. In addition, it can be seen using excessively small patch size or large shift step size on real datasets can lead to incomplete feature acquisition, thus causing overfitting and excessively large models. Therefore, it is advisable to select a moderate patch size and shift step size for PT-TPAM for training.

\subsubsection{Measuring Benchmark Performance}\label{Measuring Benchmark Performance}

In this study, we evaluate TPAM in comparison to a set of benchmark algorithms for image classification. The baseline comprises black-box models, including DNNs and CNNs. For DNNs, we set the hidden layers to $[512,2048]$. As for CNNs, we select Resnet18, which shares the same input pattern as TPAM. Furthermore, among the interpretable additive models, we choose NAM\cite{Agarwal2021Neural}, SPAM\cite{Dubey2022Scalable}, GAM\cite{Hastie1990Generalized}, and GA$^2$M\cite{Lou2013Accurate} as baselines. V-TPAM and SPAM are set to first and second-order polynomials, while T-TPAM, MPT-TPAM, and PT-TPAM are set to second-order. The rank for both TPAM and SPAM is chosen from $[1, 200]$. All self-interpretation models utilize linear functions. To evaluate the performance of TPAM on any dimensional datasets, we conducted experiments on both 2D and 3D datasets. For the 2D task, we use datasets a and b for time series classification task and MNIST and Fashion datasets for 2D image classification. The CIFAR-10 and STL-10 datasets are utilized for 3D image classification tasks.

Table \ref{tab:different_inter_xperiments} summarizes the results we have obtained. Some conclusions can be drawn from the results. We have found that TPAM performs better than previous interpretable models on all datasets, with second-order TPAM performing significantly better than the baseline of all interpretable models. Next, the number of parameters in V-TPAM and TPAM is drastically reduced, resulting in model compression. It appears that the use of various patches during the training of PT-TPAM and MPT-TPAM has resulted in improved performance. However, this has also led to an increase in parameters on MPT-TPAM. For large datasets, TPAM can perform the same expansion as SPAM, while NAM and GA$^2$M cannot fit higher-order interactions due to increased computational effort. Compared to the uninterpretable baseline, TPAM demonstrates superior performance and parameter reduction compared to DNNs. However, TPAM still exhibits a significant disparity in classification accuracy when compared to CNNs.

\subsection{Interpretation Evaluation}\label{Interpretation Evaluation}

TPAM exhibits exceptional interpretability, which is another significant advantage. We proceed to assess the self-interpretability of TPAM and the viability of P-CAM and P-CAM methods derived from it. 

\subsubsection{Self-Explanatory Performance Evaluation}\label{Self-Explanatory Performance}

We extract the feature importance of each pixel after the input passes through the model to create an interpretation map. In this section, we introduce a series of established metrics used to evaluate the interpretability of models. Then, we evaluate the explanations produced by both the TPAM model and the post-hoc explanatory model. Besides, our evaluation highlights the exceptional explanatory performance of TPAM and analysis delves into the factors contributing to the enhanced performance of TPAM, providing comprehensive explanations.

\textbf{Evaluation Metrics}. Here are four commonly used evaluation metrics:

Average Drop (AD) \cite{chattopadhay2018grad} quantifies the impact on the predicted class probability when the input image is masked using an interpretation map. Let $Y_{n}^{c}$ and $O_{n}^{c}$ denote the predicted probability of class $c$ when the $n$-th test image and its corresponding masked image are utilized as inputs to the model.  Then, AD is expressed as $\mathrm{AD}(\%):=\frac{1}{N} \sum_{n=1}^N \frac{\max \left(0, Y_n^c-O_n^c\right)}{Y_n^c} \times 100$. Here, $N$ represents the total number of test images. AD calculates the degree of predictive power reduction solely caused by masking the images, where the predictive probability of the model decreases as important pixels are removed. Lower scores indicate superior results.

Average Increase (AI) \cite{chattopadhay2018grad} quantifies the percentage of images in which the masked image yields a higher class probability compared to the original image, where the predictive probability of the model gradually increases as the number of significant pixels increases. AI is expressed as $\operatorname{AI}(\%):=\frac{1}{N} \sum_{n=1}^N \frac{\operatorname{sign}\left(Y_n^c<O_n^c\right)}{N} \times 100$. For $O_{n}^{c}$, we add the top 25\% of significant original images to the baseline image by interpretation map. Higher values indicate preferable outcomes.

Deletion (Del)\cite{Petsiuk2018RISE} and Insertion (Ins)\cite{Petsiuk2018RISE} are used to analyze the impact of individual pixels on image classification. Del involves progressively removing pixels based on their importance and observing the effect on the class prediction probability. The method generates a probability curve, where the area under this curve (Del) becomes smaller as more significant pixels are removed, indicating a good explanation if there's a steep decrease. Conversely, Ins starts with a baseline image and adds the most significant pixels incrementally. This generates a probability curve where a sharp increase and a larger area under the curve indicate a more effective explanation, demonstrating the importance of the added pixels.

\textbf{Interpretable Performance of TPAM. }TPAM exhibits strong interpretability, particularly in image classification tasks and making model decisions. It calculates the contribution value of each pixel, which is then directly summed to generate the result. With the equation \eqref{eq11}, an interpretation map of the same size as the image is created without the need for an additional interpretation model. In this section, we evaluate the interpretation maps generated by TPAM using its interpretations. For comparison, we employ additional post-hoc explanatory models, including LIME \cite{Ribeiro2016Why} and SHAP\cite{Lundberg2017Unified}, to extract decisions within the TAPM without utilizing self-interpretation. Our assessment metrics for evaluating the reliability of the interpreted maps include AD, AI, Ins, and Del. To present the interpretation map graphs and report the assessment results, we randomly selected 2000 images from the MNIST and F-MNIST datasets, respectively. The results are depicted in Table \ref{tab:LIME SHAP TPAM}.

\begin{figure}[t]
\centering
\includegraphics[width=0.9\columnwidth]{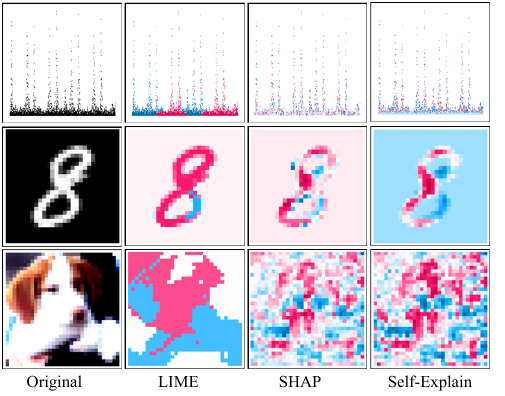} %
\caption{Interpretation maps generated by the TPAM self-interpretation and post-hoc interpretation methods (LIME and SHAP) are created for  Hearbeat, MNIST and CIFAR-10 datasets. Red represents positive contributions while blue represents negative contributions. The first row represents the model's decision against the healthy people. The second row represents the decision against the digit 8. The third row focuses on the decision against dogs. }
\label{fig:Explanation of LIME, SHAP and TPAM generated on MNIST and F-MNIST}
\end{figure}

\begin{table}[t]
	\centering
	\caption{A comparison between TPAM model self-interpretation and other post-hoc explanation methods.}
	\resizebox{\linewidth}{!}{
		\begin{tabular}{l c c c c c c c c}
   \hline \hline
			\multirow{2.5}{*}{Methods} &  \multicolumn{4}{c}{MNIST} & \multicolumn{4}{c}{Fashion-MNIST}\\
			\cmidrule(lr){2-5}
			\cmidrule(lr){6-9}
                
			&AD$\downarrow$ & AI$\uparrow$ & Ins$\uparrow$ & Del$\downarrow$ & AD$\downarrow$ & AI$\uparrow$& Ins$\uparrow$ & Del$\downarrow$ \\
			\midrule
			LIME & 9.97 & 23.04 & 0.78 & 0.15 & 48.13 & 10.90 &0.54 &0.35 \\
   			SHAP & 1.02 & 64.73 & 0.91 & 0.07 & 21.75 & 34.26 &0.73 &0.14 \\
			\rowcolor{cyan!15}
			Self-Explain & \textbf{0.84} & \textbf{67.95} & \textbf{0.92} & \textbf{0.06} & \textbf{21.29} & \textbf{37.30} &\textbf{0.75} &\textbf{0.14} \\

   \hline \hline
	\end{tabular}}
	\label{tab:LIME SHAP TPAM}
\end{table}

Fig. \ref{fig:Explanation of LIME, SHAP and TPAM generated on MNIST and F-MNIST} presents the interpretation maps generated by LIME, SHAP, and TPAM. From Table \ref{tab:LIME SHAP TPAM}, it can be observed that TPAM-produced interpretation maps outperform those generated by SHAP in all evaluation metrics, while LIME achieves worse performance. This outcome is anticipated because TPAM generates interpretation maps through self-interpretation, wherein each pixel's weight directly influences the outcome without any errors. On the other hand, SHAP employs gradient-based weight determination, which is prone to errors and may result in vanishing gradients. Furthermore, LIME relies on function approximation to obtain weight values, leading to significant approximation errors and unstable interpretation map results.

\textbf{Interpretation Performance Evaluation of Additive Models}. In study \ref{Ablation Studies}, 
we find that TPAM significantly outperforms the other baseline additive models. Here we leverage the interpretability of additive models to generate interpretation maps for TPAM, SPAM, and GAM models to ensure an understanding of the improvements achieved.

In the results depicted in Fig. \ref{fig:Explanation of GAM,TPAM,SPAM generated on MNIST and F-MNIST}, the interpretation maps generated by TPAM and GAM exhibit smoother, whereas SPAM displays a significant amount of noise. This observation is further supported by the findings presented in Table \ref{tab: The variance of explanation maps}, where the generated interpretation maps were normalized and their variances were measured. As expected, TPAM, with its ability to capture feature interactions in spatial locations, showcases lower variance and generates superior interpretation maps. In contrast, SPAM struggles to account for the spatial relationships between features due to its reliance on vector inputs.

\begin{table*}[t]
	\centering
	\caption{Comparison of P-CAM and PI-CAM with other CAM for generating saliency maps. P-CAM and PI-CAM are close to or exceed existing methods in AD, AI, Ins, and Del metrics.}
    \label{tab: Various CAM performance}
	\resizebox{\linewidth}{!}{
		\begin{tabular}{l c c c c c c c c c c c c c c c c}
   \hline \hline

   &\multicolumn{8}{c}{VGG-16}& \multicolumn{8}{c}{Inception-v3}\\
    \midrule
			\multirow{2.5}{*}{Methods} &  \multicolumn{4}{c}{Mini-ImageNet} & \multicolumn{4}{c}{CUB-200}&\multicolumn{4}{c}{Mini-ImageNet} & \multicolumn{4}{c}{CUB-200}\\
			\cmidrule(lr){2-5}
			\cmidrule(lr){6-9}
			\cmidrule(lr){10-13}
			\cmidrule(lr){14-17} 
			&AD$\downarrow$ & AI$\uparrow$ & Ins$\uparrow$ & Del$\downarrow$ & AD$\downarrow$ & AI$\uparrow$& Ins$\uparrow$ & Del$\downarrow$ & AD$\uparrow$ $\downarrow$& AI$\uparrow$ & Ins$\uparrow$ & Del$\downarrow$ & AD$\downarrow$ & AI$\uparrow$& Ins$\uparrow$ & Del$\downarrow$\\
			\midrule
Grad-CAM                                     & 44.57          & 10.93          & 68.44          & 13.32          & 20.63          & \textbf{32.59} & 68.79          & 3.77          & 48.03          & 8.77          & 72.07          & 17.28          & 30.50          & 24.97          & 64.22          & 3.50          \\
Grad-CAM++                                   & 43.92          & \textbf{11.53} & 68.57          & 14.04          & 21.24          & 30.67          & 67.84          & 3.46          & 51.37          & 8.00          & 70.98          & 17.63          & 31.89          & 23.16          & 62.93          & 3.78          \\
XGrad-CAM                                    & 45.32          & 11.00          & 68.92          & 13.16          & 20.31          & 31.44          & 68.03          & 3.47          & 46.87          & \textbf{9.10} & \textbf{72.82} & 17.22          & \textbf{29.90} & 24.92          & 64.53          & 3.50          \\
SGrad-CAM++                             & 58.75          & 6.87           & 60.46          & 21.48          & 27.00          & 24.86          & 66.02          & 4.63          & 54.73          & 5.97          & 68.48          & 21.24          & 32.56          & 21.99          & 63.60          & 4.01          \\
Layer-CAM                                    & 48.01          & 9.97           & 66.74          & 14.00          & 21.44          & 29.73          & 68.27          & 3.45          & 51.33          & 7.17          & 70.90          & 17.97          & 30.50          & 23.76          & 64.02          & 3.57          \\
Score-CAM                                    & 48.79          & 9.63           & 66.07          & 15.61          & 21.49          & 29.73          & 66.44          & 3.68          & 50.87          & 8.00          & 69.34          & 19.30          & 30.49          & 23.79          & 63.16          & 3.84          \\
\rowcolor{cyan!7}
P-CAM                                        & \textbf{43.86} & 11.57          & 69.20          & 13.52          & \textbf{19.45} & 31.31          & 68.49          & 3.44          & 48.91          & 7.57          & 72.11          & 17.40          & 31.33          & \textbf{25.39} & 64.50          & \textbf{3.40} \\
\rowcolor{cyan!15}
PI-CAM                                       & 49.41          & 10.83          & \textbf{70.11} & \textbf{12.82} & 21.62          & 29.30          & \textbf{69.15} & \textbf{3.32} & \textbf{46.17} & 8.45          & 72.68          & \textbf{17.11} & 30.20          & 24.42          & \textbf{64.69} & 3.41\\
\hline\hline
\end{tabular}}
\end{table*}

\begin{figure}[t]
\centering
\includegraphics[width=0.9\columnwidth]{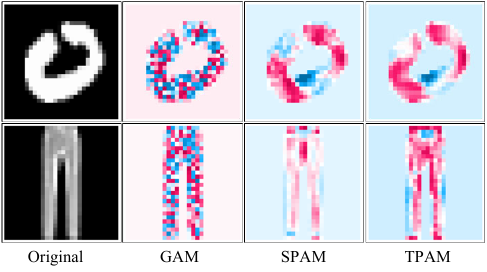} 
\caption{Interpretation maps generated by TPAM, GAM, and SPAM for MNIST and F-MNIST datasets.}

\label{fig:Explanation of GAM,TPAM,SPAM generated on MNIST and F-MNIST}
\end{figure}

\begin{table}[t]
	\centering
	\caption{Variance of interpretation maps produced by TPAM, GAM, and SPAM on MNIST and F-MNIST determines their smoothness and focus. Smaller variances indicate smoother and more focused interpretation maps.}
    \label{tab: The variance of explanation maps}
\begin{tabular}{lcc}
\hline\hline
\multicolumn{3}{c}{Variance} \\
\hline
 Model      & MNIST    & F-MNIST  \\
       \hline
GAM    & 0.00907  & 0.00927  \\
SPAM   & 0.02206  & 0.01097  \\
\rowcolor{cyan!7} TPAM   & \textbf{0.00891}  & \textbf{0.00803}  \\
\hline\hline
\end{tabular}
\end{table}

\subsubsection{Performance of P-CAM and PI-CAM}

\begin{figure*}[t]
\centering
\includegraphics[width=0.95\textwidth]{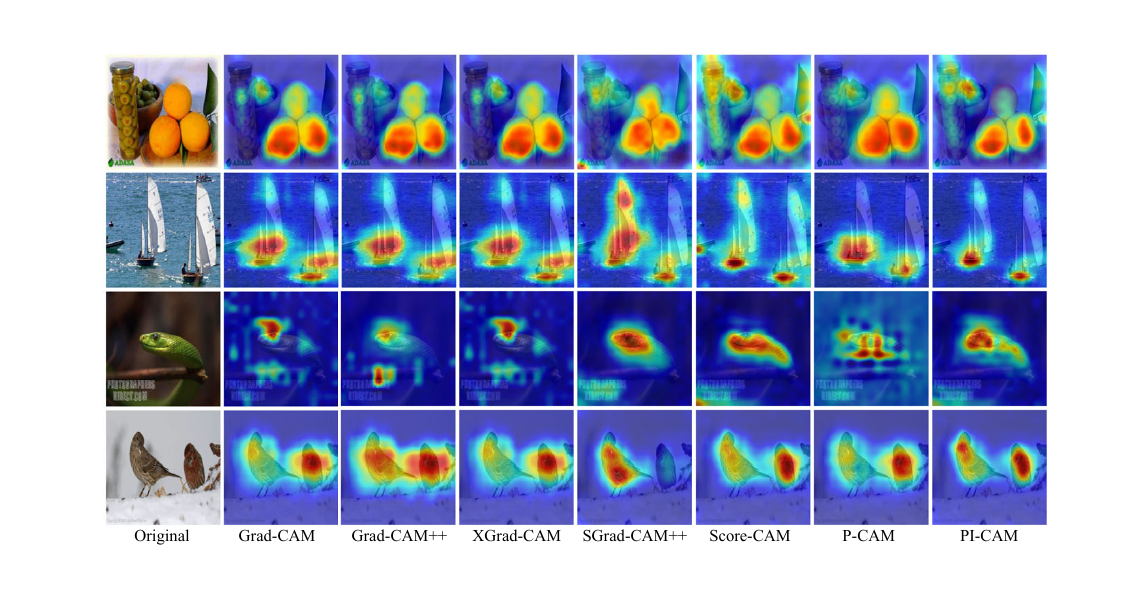}
\caption{Comparison of saliency maps generated by Grad-CAM, Grad-CAM++, XGrad-CAM, SGrad-CAM++, Layer-CAM, Score-CAM, P-CAM, and P-CAM on the VGG-16 model in Mini-Imagenet dataset.}
\label{fig:Sample saliency maps generated}
\end{figure*}

We evaluated the accuracy of P-CAM and PI-CAM in generating interpretation maps for an image classification task and assessed the effectiveness of the resulting saliency maps for image localization. The experiments utilize the VGG-16 model\cite{Simonyan2015Very} and Inception-V3 model\cite{szegedy2016rethinking}, modified as described in \cite{Zhou2016Learning}, to generate feature maps. Two publicly accessible object classification datasets, Mini-Imagenet\cite{vinyals2016matching} and CUB-200\cite{wah2011caltech}, are used in our experiments. Prior to inputting the images into the VGG-16 network, we resize them to $224\times224\times3$. Similarly, for Inception-V3, the images are resized to $299\times299\times3$. The images are normalized using the mean vector $[0.485, 0.456, 0.406]$ and the standard deviation vector $[0.229, 0.224, 0.225]$ to ensure they fall within the range of [0, 1]. Additionally, we set the initial input $\mathcal{X}_b$ for PI-CAM as an all-one tensor. Finally, we compare the saliency maps generated by Grad-CAM\cite{Selvaraju2017GradCAM}, Grad-CAM++\cite{chattopadhay2018grad}, XGrad-CAM\cite{fu2020axiom}, SGrad-CAM++\cite{Omeiza2019Smooth}, Layer-CAM\cite{jiang2021layercam}, and Score-CAM\cite{Wang2020ScoreCAM}.

\textbf{Faithfulness Evaluation. }We assess the confidence of PI-CAM and P-CAM by conducting quantitative testing on 2000 images from each of the two datasets. We measured the performance using the AD, AI, Ins, and Del metrics.

The evaluation results of PI-CAM and P-CAM, along with other CAM methods, are presented in Table \ref{tab: Various CAM performance}. The results demonstrate that PI-CAM and P-CAM exhibit superior performance on multiple metrics for both network structures and datasets. P-CAM performs well on the AD and AI metrics, which can be attributed to its emphasis on highlighting significant regions in the saliency maps through weight allocation. In contrast, PI-CAM outperforms other CAM methods on the Ins and Del metrics. This advantage can be attributed to PI-CAM's ability to generate saliency maps with reduced noise and enhanced focus on important network regions. 

Notably, the interference of noise affects the evaluation of saliency maps by the network, rendering existing CAM methods sensitive to noise and unable to handle feature maps containing noise. Additionally, the performance of the remaining CAM methods is comparably close, with a noticeable gap separating their performance from that of P-CAM and PI-CAM in smaller interpretation task classes. Fig.  \ref{fig:Sample saliency maps generated} presents sample saliency maps generated by PI-CAM, P-CAM, and other CAM. Our method stands out by producing clearer and less noisy saliency maps, displaying more concentrated labeled regions.

\textbf{Localization Evaluation. }We conduct an evaluation of the localization capability, which is another important metric used to assess the performance of saliency maps. Instead of the conventional approach of locating the maximum value of the saliency map, we employ the Proportion to measure the energy of the saliency map. This metric evaluates the saliency map's ability to accurately identify the number of salient regions within the bounding box, thereby examining the distribution of noise in the saliency map. To perform quantitative analysis, we select 2000 images from the CUB-200 dataset. The results of this analysis are presented in Table \ref{tab:Location capability}.

Table \ref{tab:Location capability} shows that PI-CAM concentrates over 80\% of its energy within the true bounding box in the saliency maps, while P-CAM and other CAM methods allocate approximately 70\% of the energy within the true bounding box. This observation suggests that the saliency map generated by PI-CAM is cleaner and exhibits a higher concentration of energy.

\begin{table}[t]
	\centering
	\caption{Comparison of P-CAM and PI-CAM with other CAM on the localization task. Higher proportions are associated with better localization performance, and PI-CAM far outperforms existing methods on the proportions metric.}
    \label{tab:Location capability}
\begin{tabular}{l c c}
\hline\hline
\multirow{2}{*}{Method} & \multicolumn{2}{c}{Proportion}       \\
\cmidrule(lr){2-3}
                        & VGG-16          & Inception-V3    \\
\midrule
Grad-CAM                & 71.247          & 72.304          \\
Grad-CAM++              & 70.623          & 72.648          \\
XGrad-CAM               & 71.281          & 72.491          \\
SGrad-CAM++             & 71.907          & 70.848          \\
Layer-CAM               & 72.067          & 75.122          \\
Score-CAM               & 71.380          & 73.375          \\
\rowcolor{cyan!7}
P-CAM                   & 70.494          & 72.305          \\
\rowcolor{cyan!15}
PI-CAM                  & \textbf{81.817} & \textbf{78.195} \\
\hline\hline
\end{tabular}
\end{table}

\section{Conclusion}

In this paper, we propose TPAM, an innovative interpretable polynomial model utilizing higher-order tensor inputs, 
bypassing potential performance losses associated with traditional vectorization. To effectively learn interactions between higher-order features, we adopt a hierarchical low-order symmetric tensor approximation in weight space, reducing computational complexity. Compared to existing additive models, TPAM offers considerable advantages in parameter efficiency and accuracy performance. Additionally, its decision-making process is transparent, enabling the extraction of each feature's importance for credible interpretations. Moreover, we apply TPAM to the classification header in CNNs and accordingly propose two new algorithms for CAM generation. These algorithms directly process feature maps, applying a factorization weighting tensor matching the feature maps' size to create more comprehensive saliency maps.

\bibliographystyle{IEEEtran}
\bibliography{TPAM}

\begin{thebibliography}{10}
\providecommand{\url}[1]{#1}
\csname url@samestyle\endcsname
\providecommand{\newblock}{\relax}
\providecommand{\bibinfo}[2]{#2}
\providecommand{\BIBentrySTDinterwordspacing}{\spaceskip=0pt\relax}
\providecommand{\BIBentryALTinterwordstretchfactor}{4}
\providecommand{\BIBentryALTinterwordspacing}{\spaceskip=\fontdimen2\font plus
\BIBentryALTinterwordstretchfactor\fontdimen3\font minus \fontdimen4\font\relax}
\providecommand{\BIBforeignlanguage}[2]{{%
\expandafter\ifx\csname l@#1\endcsname\relax
\typeout{** WARNING: IEEEtran.bst: No hyphenation pattern has been}%
\typeout{** loaded for the language `#1'. Using the pattern for}%
\typeout{** the default language instead.}%
\else
\language=\csname l@#1\endcsname
\fi
#2}}
\providecommand{\BIBdecl}{\relax}
\BIBdecl

\bibitem{Guidotti2019Survey}
R.~Guidotti, A.~Monreale, S.~Ruggieri, F.~Turini, F.~Giannotti, and D.~Pedreschi, ``A survey of methods for explaining black box models,'' \emph{ACM computing surveys (CSUR)}, vol.~51, no.~5, pp. 1--42, 2018.

\bibitem{Bodria2023Benchmarking}
F.~Bodria, F.~Giannotti, R.~Guidotti, F.~Naretto, D.~Pedreschi, and S.~Rinzivillo, ``Benchmarking and survey of explanation methods for black box models,'' \emph{Data Mining and Knowledge Discovery}, pp. 1--60, 2023.

\bibitem{Chang2022NODEGAM}
C.-H. Chang, R.~Caruana, and A.~Goldenberg, ``Node-gam: Neural generalized additive model for interpretable deep learning,'' \emph{arXiv preprint arXiv:2106.01613}, 2021.

\bibitem{Li2022Interpretable}
X.~Li, H.~Xiong, X.~Li, X.~Wu, X.~Zhang, J.~Liu, J.~Bian, and D.~Dou, ``Interpretable deep learning: Interpretation, interpretability, trustworthiness, and beyond,'' \emph{Knowledge and Information Systems}, vol.~64, no.~12, pp. 3197--3234, 2022.

\bibitem{Lundberg2017Unified}
S.~M. Lundberg and S.-I. Lee, ``A unified approach to interpreting model predictions,'' \emph{Advances in neural information processing systems}, vol.~30, 2017.

\bibitem{Petsiuk2018RISE}
V.~Petsiuk, A.~Das, and K.~Saenko, ``Rise: Randomized input sampling for explanation of black-box models,'' \emph{arXiv preprint arXiv:1806.07421}, 2018.

\bibitem{Ribeiro2016Why}
M.~T. Ribeiro, S.~Singh, and C.~Guestrin, ``" why should i trust you?" explaining the predictions of any classifier,'' in \emph{Proceedings of the 22nd ACM SIGKDD international conference on knowledge discovery and data mining}, 2016, pp. 1135--1144.

\bibitem{Puri2021CoFrNets}
I.~Puri, A.~Dhurandhar, T.~Pedapati, K.~Shanmugam, D.~Wei, and K.~R. Varshney, ``Cofrnets: interpretable neural architecture inspired by continued fractions,'' \emph{Advances in neural information processing systems}, vol.~34, pp. 21\,668--21\,680, 2021.

\bibitem{Fel2023Don}
T.~Fel, M.~Ducoffe, D.~Vigouroux, R.~Cad{\`e}ne, M.~Capelle, C.~Nicod{\`e}me, and T.~Serre, ``Don't lie to me! robust and efficient explainability with verified perturbation analysis,'' in \emph{Proceedings of the IEEE/CVF Conference on Computer Vision and Pattern Recognition}, 2023, pp. 16\,153--16\,163.

\bibitem{Hastie1990Generalized}
T.~J. Hastie, ``Generalized additive models,'' in \emph{Statistical models in S}.\hskip 1em plus 0.5em minus 0.4em\relax Routledge, 2017, pp. 249--307.

\bibitem{Zhuang2020Interpretable}
H.~Zhuang, X.~Wang, M.~Bendersky, A.~Grushetsky, Y.~Wu, P.~Mitrichev, E.~Sterling, N.~Bell, W.~Ravina, and H.~Qian, ``Interpretable learning-to-rank with generalized additive models,'' \emph{arXiv preprint arXiv:2005.02553}, 2020.

\bibitem{Lou2013Accurate}
Y.~Lou, R.~Caruana, J.~Gehrke, and G.~Hooker, ``Accurate intelligible models with pairwise interactions,'' in \emph{Proceedings of the 19th ACM SIGKDD international conference on Knowledge discovery and data mining}, 2013, pp. 623--631.

\bibitem{Agarwal2021Neural}
R.~Agarwal, L.~Melnick, N.~Frosst, X.~Zhang, B.~Lengerich, R.~Caruana, and G.~E. Hinton, ``Neural additive models: Interpretable machine learning with neural nets,'' \emph{Advances in neural information processing systems}, vol.~34, pp. 4699--4711, 2021.

\bibitem{Dubey2022Scalable}
A.~Dubey, F.~Radenovic, and D.~Mahajan, ``Scalable interpretability via polynomials,'' \emph{Advances in neural information processing systems}, vol.~35, pp. 36\,748--36\,761, 2022.

\bibitem{Nie2017Generating}
J.~Nie, ``Generating polynomials and symmetric tensor decompositions,'' \emph{Foundations of Computational Mathematics}, vol.~17, pp. 423--465, 2017.

\bibitem{Brachat2010Symmetric}
J.~Brachat, P.~Comon, B.~Mourrain, and E.~Tsigaridas, ``Symmetric tensor decomposition,'' \emph{Linear Algebra and its Applications}, vol. 433, no. 11-12, pp. 1851--1872, 2010.

\bibitem{Selvaraju2017GradCAM}
R.~R. Selvaraju, M.~Cogswell, A.~Das, R.~Vedantam, D.~Parikh, and D.~Batra, ``Grad-cam: Visual explanations from deep networks via gradient-based localization,'' in \emph{Proceedings of the IEEE international conference on computer vision}, 2017, pp. 618--626.

\bibitem{Omeiza2019Smooth}
D.~Omeiza, S.~Speakman, C.~Cintas, and K.~Weldermariam, ``Smooth grad-cam++: An enhanced inference level visualization technique for deep convolutional neural network models,'' \emph{arXiv preprint arXiv:1908.01224}, 2019.

\bibitem{chattopadhay2018grad}
A.~Chattopadhay, A.~Sarkar, P.~Howlader, and V.~N. Balasubramanian, ``Grad-cam++: Generalized gradient-based visual explanations for deep convolutional networks,'' in \emph{2018 IEEE winter conference on applications of computer vision (WACV)}.\hskip 1em plus 0.5em minus 0.4em\relax IEEE, 2018, pp. 839--847.

\bibitem{zeiler2014visualizing}
M.~D. Zeiler and R.~Fergus, ``Visualizing and understanding convolutional networks,'' in \emph{Computer Vision--ECCV 2014: 13th European Conference, Zurich, Switzerland, September 6-12, 2014, Proceedings, Part I 13}.\hskip 1em plus 0.5em minus 0.4em\relax Springer, 2014, pp. 818--833.

\bibitem{springenberg2014striving}
J.~T. Springenberg, A.~Dosovitskiy, T.~Brox, and M.~Riedmiller, ``Striving for simplicity: The all convolutional net,'' \emph{arXiv preprint arXiv:1412.6806}, 2014.

\bibitem{Desai2020AblationCAM}
H.~G. Ramaswamy \emph{et~al.}, ``Ablation-cam: Visual explanations for deep convolutional network via gradient-free localization,'' in \emph{proceedings of the IEEE/CVF winter conference on applications of computer vision}, 2020, pp. 983--991.

\bibitem{Dabkowski2017Real}
P.~Dabkowski and Y.~Gal, ``Real time image saliency for black box classifiers,'' \emph{Advances in neural information processing systems}, vol.~30, 2017.

\bibitem{wang2020ss}
H.~Wang, R.~Naidu, J.~Michael, and S.~S. Kundu, ``Ss-cam: Smoothed score-cam for sharper visual feature localization,'' \emph{arXiv preprint arXiv:2006.14255}, 2020.

\bibitem{Fong2019Understanding}
R.~Fong, M.~Patrick, and A.~Vedaldi, ``Understanding deep networks via extremal perturbations and smooth masks,'' in \emph{Proceedings of the IEEE/CVF international conference on computer vision}, 2019, pp. 2950--2958.

\bibitem{Fong2017Interpretable}
R.~C. Fong and A.~Vedaldi, ``Interpretable explanations of black boxes by meaningful perturbation,'' in \emph{Proceedings of the IEEE international conference on computer vision}, 2017, pp. 3429--3437.

\bibitem{Schulz2020Restricting}
K.~Schulz, L.~Sixt, F.~Tombari, and T.~Landgraf, ``Restricting the flow: Information bottlenecks for attribution,'' \emph{arXiv preprint arXiv:2001.00396}, 2020.

\bibitem{Wang2020ScoreCAM}
H.~Wang, Z.~Wang, M.~Du, F.~Yang, Z.~Zhang, S.~Ding, P.~Mardziel, and X.~Hu, ``Score-cam: Score-weighted visual explanations for convolutional neural networks,'' in \emph{Proceedings of the IEEE/CVF conference on computer vision and pattern recognition workshops}, 2020, pp. 24--25.

\bibitem{BanyMuhammad2020EigenCAM}
M.~B. Muhammad and M.~Yeasin, ``Eigen-cam: Class activation map using principal components,'' in \emph{2020 international joint conference on neural networks (IJCNN)}.\hskip 1em plus 0.5em minus 0.4em\relax IEEE, 2020, pp. 1--7.

\bibitem{Zhou2016Learning}
B.~Zhou, A.~Khosla, A.~Lapedriza, A.~Oliva, and A.~Torralba, ``Learning deep features for discriminative localization,'' in \emph{Proceedings of the IEEE conference on computer vision and pattern recognition}, 2016, pp. 2921--2929.

\bibitem{fu2020axiom}
R.~Fu, Q.~Hu, X.~Dong, Y.~Guo, Y.~Gao, and B.~Li, ``Axiom-based grad-cam: Towards accurate visualization and explanation of cnns,'' \emph{arXiv preprint arXiv:2008.02312}, 2020.

\bibitem{jiang2021layercam}
P.-T. Jiang, C.-B. Zhang, Q.~Hou, M.-M. Cheng, and Y.~Wei, ``Layercam: Exploring hierarchical class activation maps for localization,'' \emph{IEEE Transactions on Image Processing}, vol.~30, pp. 5875--5888, 2021.

\bibitem{naidu2020cam}
R.~Naidu, A.~Ghosh, Y.~Maurya, S.~S. Kundu \emph{et~al.}, ``Is-cam: Integrated score-cam for axiomatic-based explanations,'' \emph{arXiv preprint arXiv:2010.03023}, 2020.

\bibitem{Shin1991pisigma}
Y.~Shin and J.~Ghosh, ``The pi-sigma network: An efficient higher-order neural network for pattern classification and function approximation,'' in \emph{IJCNN-91-Seattle international joint conference on neural networks}, vol.~1.\hskip 1em plus 0.5em minus 0.4em\relax IEEE, 1991, pp. 13--18.

\bibitem{Li2003SigmaPiSigma}
C.-K. Li, ``A sigma-pi-sigma neural network (spsnn),'' \emph{Neural Processing Letters}, vol.~17, pp. 1--19, 2003.

\bibitem{Simonyan2015Very}
K.~Simonyan and A.~Zisserman, ``Very deep convolutional networks for large-scale image recognition,'' \emph{arXiv preprint arXiv:1409.1556}, 2014.

\bibitem{voutriaridis2003ridge}
C.~Voutriaridis, Y.~S. Boutalis, and B.~G. Mertzios, ``Ridge polynomial networks in pattern recognition,'' in \emph{Proceedings EC-VIP-MC 2003. 4th EURASIP Conference focused on Video/Image Processing and Multimedia Communications (IEEE Cat. No. 03EX667)}, vol.~2.\hskip 1em plus 0.5em minus 0.4em\relax IEEE, 2003, pp. 519--524.

\bibitem{oh2003polynomial}
S.-K. Oh, W.~Pedrycz, and B.-J. Park, ``Polynomial neural networks architecture: analysis and design,'' \emph{Computers \& Electrical Engineering}, vol.~29, no.~6, pp. 703--725, 2003.

\bibitem{Chrysos2020Pnets}
G.~G. Chrysos, S.~Moschoglou, G.~Bouritsas, Y.~Panagakis, J.~Deng, and S.~Zafeiriou, ``P-nets: Deep polynomial neural networks,'' in \emph{Proceedings of the IEEE/CVF Conference on Computer Vision and Pattern Recognition}, 2020, pp. 7325--7335.

\bibitem{Chrysos2022Deep}
G.~G. Chrysos, S.~Moschoglou, G.~Bouritsas, J.~Deng, Y.~Panagakis, and S.~Zafeiriou, ``Deep polynomial neural networks,'' \emph{IEEE transactions on pattern analysis and machine intelligence}, vol.~44, no.~8, pp. 4021--4034, 2021.

\bibitem{Chrysos2023Regularization}
G.~G. Chrysos, B.~Wang, J.~Deng, and V.~Cevher, ``Regularization of polynomial networks for image recognition,'' in \emph{Proceedings of the IEEE/CVF Conference on Computer Vision and Pattern Recognition}, 2023, pp. 16\,123--16\,132.

\bibitem{glorot2010understanding}
X.~Glorot and Y.~Bengio, ``Understanding the difficulty of training deep feedforward neural networks,'' in \emph{Proceedings of the thirteenth international conference on artificial intelligence and statistics}.\hskip 1em plus 0.5em minus 0.4em\relax JMLR Workshop and Conference Proceedings, 2010, pp. 249--256.

\bibitem{he2015delving}
K.~He, X.~Zhang, S.~Ren, and J.~Sun, ``Delving deep into rectifiers: Surpassing human-level performance on imagenet classification,'' in \emph{Proceedings of the IEEE international conference on computer vision}, 2015, pp. 1026--1034.

\bibitem{xiao2017fashion}
H.~Xiao, K.~Rasul, and R.~Vollgraf, ``Fashion-mnist: a novel image dataset for benchmarking machine learning algorithms,'' \emph{arXiv preprint arXiv:1708.07747}, 2017.

\bibitem{lecun2015deep}
Y.~LeCun, Y.~Bengio, and G.~Hinton, ``Deep learning,'' \emph{nature}, vol. 521, no. 7553, pp. 436--444, 2015.

\bibitem{krizhevsky2009learning}
A.~Krizhevsky, G.~Hinton \emph{et~al.}, ``Learning multiple layers of features from tiny images,'' 2009.

\bibitem{kingma2014adam}
D.~P. Kingma and J.~Ba, ``Adam: A method for stochastic optimization,'' \emph{arXiv preprint arXiv:1412.6980}, 2014.

\bibitem{pennington2017resurrecting}
J.~Pennington, S.~Schoenholz, and S.~Ganguli, ``Resurrecting the sigmoid in deep learning through dynamical isometry: theory and practice,'' \emph{Advances in neural information processing systems}, vol.~30, 2017.

\bibitem{szegedy2016rethinking}
C.~Szegedy, V.~Vanhoucke, S.~Ioffe, J.~Shlens, and Z.~Wojna, ``Rethinking the inception architecture for computer vision,'' in \emph{Proceedings of the IEEE conference on computer vision and pattern recognition}, 2016, pp. 2818--2826.

\bibitem{vinyals2016matching}
O.~Vinyals, C.~Blundell, T.~Lillicrap, D.~Wierstra \emph{et~al.}, ``Matching networks for one shot learning,'' \emph{Advances in neural information processing systems}, vol.~29, 2016.

\bibitem{wah2011caltech}
C.~Wah, S.~Branson, P.~Welinder, P.~Perona, and S.~Belongie, ``The caltech-ucsd birds-200-2011 dataset,'' 2011.

\end{thebibliography}

\end{document}